\documentclass{article}
\usepackage{spconf,amsmath,graphicx}
\usepackage{amssymb}
\usepackage{url}
\pdfoutput=1
\usepackage{multirow}
\usepackage{booktabs}

\title{FF2: A Feature Fusion two-stream framework for Punctuation Restoration}
\name{Yangjun Wu, Kebin Fang, Yao Zhao, Hao Zhang, Lifeng Shi, Mengqi zhang}
\address{Institute of Computing Innovation, Zhejiang University}
\begin{document}
%
\maketitle
\begin{abstract}
To accomplish punctuation restoration, most existing methods focus on introducing extra information (e.g., part-of-speech) or addressing the class imbalance problem. Recently, large-scale transformer-based pre-trained language models (PLMS) have been utilized widely and obtained remarkable success. However, the PLMS are trained on the large dataset with marks, which may not fit well with the small dataset without marks, causing the convergence to be not ideal. In this study, we propose a \textbf{F}eature \textbf{F}usion two-stream framework (FF2) to bridge the gap. Specifically, one stream leverages a pre-trained language model to capture the semantic feature, while another auxiliary module captures the feature at hand. We also modify the computation of multi-head attention to encourage communication among heads. Then, two features with different perspectives are aggregated to fuse information and enhance context awareness. Without additional data, the experimental results on the popular benchmark IWSLT demonstrate that FF2 achieves new SOTA performance, which verifies that our approach is effective.
\end{abstract}
\section{Introduction}
\label{sec:intro}

Punctuation restoration is a significant post-processing step in automatic speech recognition (ASR) systems because punctuation marks are not usually predicted. It can enhance the readability of speech transcripts and contribute to downstream tasks, such as machine translation, intent detection, or slot filling in dialogue systems. Thus, this task has attracted a large amount of interest.

Generally, current works could be categorized into three groups: 1) The first line \cite{klejch2017sequence,8682260,8545470} treats this problem as a machine translation task, which feeds a non-punctuation sequence and yields the output with marks. 2) Second, some studied \cite{Adversarial, Lin2020JointPO,alam-etal-2020-punctuation,shi21_interspeech} regard it as a sequence labeling task, where a punctuation mark is assigned to each word by probability. 3) The others \cite{Che2016PunctuationPF} employ a classifier to forecast a tag for each token via taking it as a token-level classification task.

To address this problem, transformer-based pre-trained language models (PLMS) have been widely applied to enhance their effectiveness. However, We argue that PLMS trained on the large-scale dataset with marks would be ineffective while fine-tuning on the dataset without marks. The main reason is that the absence of punctuation between words in a sentence may severely damage the semantics. Besides, the experiments \cite{Bottleneck, Talking} indicate that the limitation called \textit{Low-Rank Bottleneck} exists in the self-attention mechanism. Concretely, with the fixed vector size of multi-head self-attention vector\cite{attention}, increasing the number of self-attention heads would decrease the vector size in each head, which causes significant degradation of its comprehension.

Inspired by these observations, we propose a \textbf{F}eature \textbf{F}usion framework based on two-stream attentions (FF2) \footnote{The code, dataset, and evaluation results are public available at \url{https://github.com/qinqinqaq/FF2_punc_restore.git}} to mitigate the shortage. Specifically, one stream utilizes a pre-trained language model to capture the semantic feature of the sequence, and another tiny module is randomly initialized, which captures the feature information on the current dataset without punctuation. Furthermore, we also modify the computation of multi-head attention to encourage communication among heads. By the mutual benefit of the two-stream modules, we first obtain two-type feature representations. Then, the two vectors are aggregated to fuse features and  advance context awareness. Finally, we yield token-level punctuation tags as the output. Our main contributions are summarized as follows:
\begin{itemize}
\item We present a novel framework (FF2) to encourage message sharing and capture the features with different perspectives to advance the shortage of vanilla attention mechanisms. 
\item Without extra data, the results on the popular benchmark IWSLT indicate that FF2 can leverage the dataset itself and achieve the new state-of-the-art performance, demonstrating that FF2 is effective.
\item We introduce a novel computation of multi-head attention to encourage communication among heads. The ablation studies show that it can increase the expression capability of attention heads.
\end{itemize}

\section{Problem Definition}
\label{sec:Problem}
Given the input sequence $ X = (x_1, x_2, ... x_n) $ and punctuation tags $Y = (y_1, y_2,... y_n)$, punctuation restoration is defined as a token-level classification task that outputs a sequence  $ \hat{Y} = (\hat{y}_1, \hat{y}_2, ... \hat{y}_n) $ in \textit{[O, COMMA, PERIOD, QUESTION]}, the \textit{O} denotes the label is None, where n is the length of the input utterance. 

\section{Methodology}
\begin{figure}[t]
  \centering
  \includegraphics[width=\linewidth]{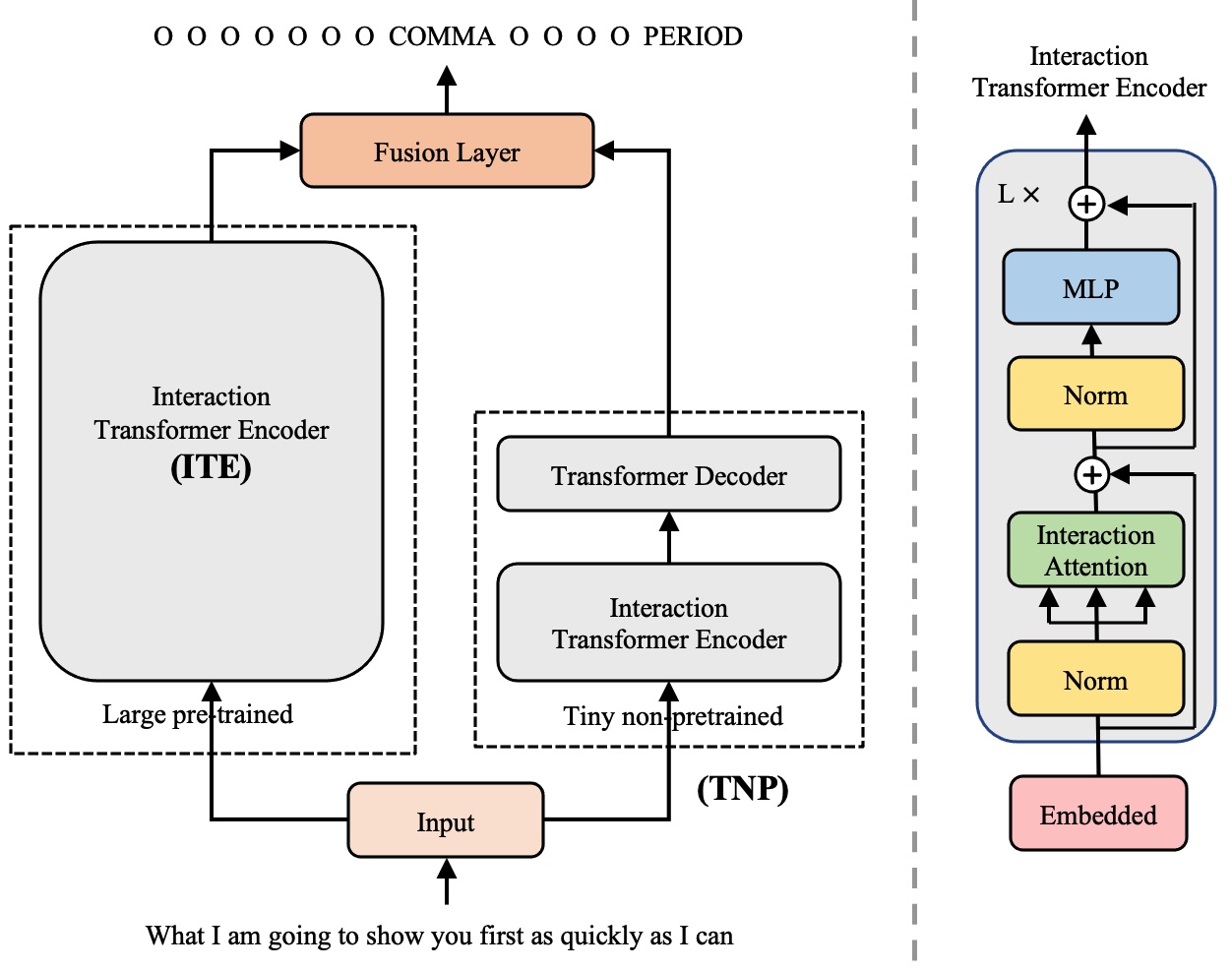}
  \caption{ Overall architecture. L refers to the number of layers. }
  \label{fig:ff2}
\end{figure}

In this section, we will describe the modification of self-attention and FF2 in detail. As described in Figure \ref{fig:ff2}, FF2 contains three core components: Interaction Transformer Encoder module (ITE), Tiny Non-pretrained module (TNP), and Fusion Layer (FL).

\subsection{Multi-head self-attention (Interaction)}
\begin{figure}[t]
  \includegraphics[width=\linewidth]{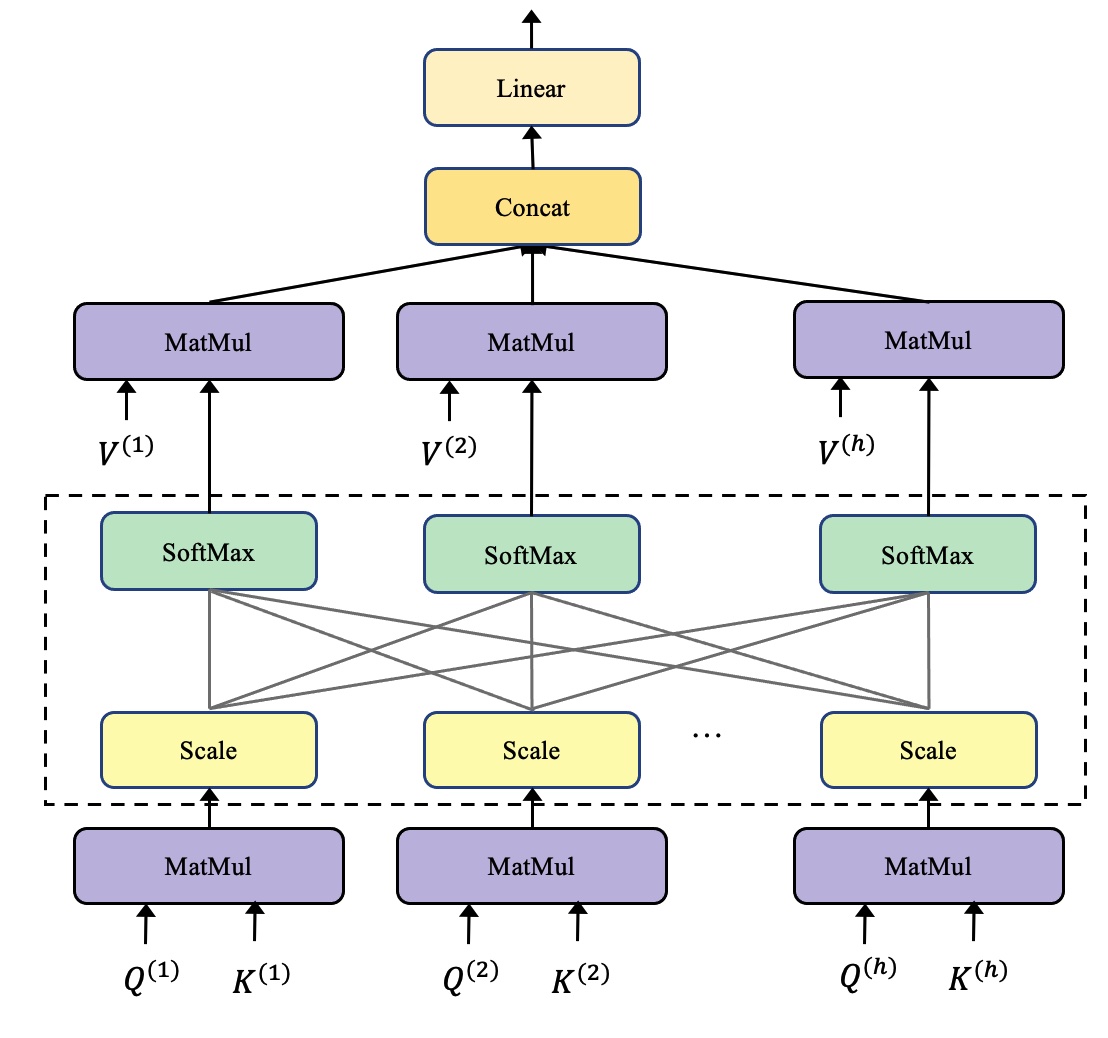}
  \caption{The calculation of interaction self-attention between Q, K, and V in heads.}
  \label{fig:attention}
\end{figure}

For the conventional self-attention\cite{attention} mechanism, the computation of scaled dot-product lies:

\begin{gather}
    Attn^{(k)}{(Q,K,V)} = softmax(\frac{J^{(k)}}{\sqrt{d_{emb} / H}})V^{(k)} \label{equa.1} \\
    J^{(k)} = {Q^{(k)}K^{(k)}}^\mathrm{T} \\
    Q = x_i W_Q^h, K = x_j W_K^h, V = x_j W_V^h\label{equa1}
\end{gather}

Here, $J^{(k)}$ denotes the dot-product between query $Q$ and key $K$ for the {$k$-th} head, $d_{emb}$ refers to the embedding size, $H$ is the head numbers. 

To alleviate the inherent issue (\textit{Low-Rank Bottleneck}) in multi-head self-attention, we modify the computation of self-attention as a two-stage procedure. First, the attention scores $J^{(k)}$ are obtained via equation \ref{equa.1}. Then, these scores are aggregated to update the attention weights via a projection component $\phi$. 
For the component $\phi$, we introduce a new matrix $P_\lambda$ \footnote{$P_\lambda$ is initialized with standard deviations of $0.1/\sqrt{d_{emb}}$} to share information with sibling heads. Then, we update the $J^{(k)}$ as $\hat{J^{(k)}}$ to represent the new attention weights as below:

\begin{scriptsize}
\begin{gather}
\begin{pmatrix}
\hat{J^{(1)}} \\\\
\hat{J^{(2)}} \\\\
\vdots \\\\
\hat{J^{(h)}} \\\\
\end{pmatrix} = P_\lambda\begin{pmatrix}
J^{(1)} \\\\
J^{(2)} \\\\
\vdots \\\\
J^{(h)} \\\\
\end{pmatrix} + 
\begin{pmatrix}
J^{(1)} \\\\
J^{(2)} \\\\
\vdots \\\\
J^{(h)} \\\\
\end{pmatrix},
P_\lambda = 
\begin{pmatrix}
    \lambda_{11} & \cdots & \lambda_{1h}\\\\
    \lambda_{21} & \cdots & \lambda_{2h}\\\\
    \vdots & \ddots &  \vdots \\\\
    \lambda_{h1} & \cdots &  \lambda_{hh}\\\\
\end{pmatrix}
\end{gather}
\end{scriptsize}

Here, $\lambda_{hh}$ denotes the learned parameter. We modify the Equation \ref{equa.1} to novel Equation \ref{equa5} as the interaction scores:
\begin{equation}
    Attn^{(k)}{(Q,K,V)} = softmax(\frac{\hat{J^{(k)}}}{\sqrt{d_{emb} / H}})V^{(k)}\label{equa5}
\end{equation}
Second, we repeat $h$ times the projections in different $Q, K, V $, and concatenate the embeddings. Then, a linear layer is employed to obtain the final matrices $M$:
\begin{footnotesize}
\begin{equation}
 M\left(Q,K,V\right)=Concat\left({\rm Attn}^{(1)},\ldots,{\rm Attn}^{(h)}\right)W +B
\end{equation}
\end{footnotesize}Here, $W$ is a learnable parameter and $B$ is the bias. After that, we follow the standard transformer workflow and feed the matrices into the layer normalization and feed-forward layer.

\begin{table*}[!ht]
  \caption{Results in terms of precision (P \%), recall (R \%), and F1-score (F1 \%) on the English IWSLT2011 test set. We collect the results from the original papers without modification, and the highest numbers are in bold.}
  \label{tab:1}
  \centering
  \setlength{\tabcolsep}{2.5mm}{
    \begin{tabular}{ c|r|r|r|r|r|r|r|r|r|r|r|r}
       \toprule
        \multirow{2}*{\textbf{Model}} & \multicolumn{3}{c|}{\textbf{Comma}} &\multicolumn{3}{c|}{\textbf{Period}} &\multicolumn{3}{c|}{\textbf{Question}} &\multicolumn{3}{c}{\textbf{Overall}} \\
        \cline{2-13} & P & R& F1 & P & R& F1 & P & R& F1 & P & R& F1\\
        \hline
        BLSTM-CRF &58.9&59.1&59.0&68.9&72.1&70.5&71.8&60.6&65.7&66.5&63.9&65.1 \\
        Teacher-Ensemble&66.2&59.9&62.9&75.1 &73.7& 74.4&72.3&63.8&67.8&71.2&65.8&68.4\\
        DRNN-LWMA-pre& 62.9&60.8&61.9&77.3&73.7&75.5&69.6&69.6&69.6&69.9&67.2&68.6 \\
        Self-attention-word-speech& 67.4&61.1&64.1&82.5&77.4&79.9&80.1&70.2&74.8&76.7&69.6&72.9\\
        CT-Transformer &68.8&69.8&69.3&78.4&82.1&80.2&76.0&82.6&79.2&73.7&76.0&74.9\\
        SAPR & 57.2&50.8&55.9&\textbf{96.7}&\textbf{97.3}&\textbf{96.8}&70.6&69.2&70.3&78.2&74.4&77.4\\ 
        BERT-base+Adversarial&76.2&71.2&73.6&87.3&81.1&84.1&79.1&72.7&75.8&80.9&75.0&77.8\\
        BERT-large+Transfer&70.8&74.3&72.5&84.9&83.3&84.1&82.7&93.5&87.8&79.5&83.7&81.4\\
        BERT-base+FocalLoss&74.4&77.1&75.7&87.9&88.2&88.1&74.2&88.5&80.7&78.8&84.6&81.6\\
        RoBERTa-large+augmentation&76.8&76.6&76.7&88.6&89.2&88.9&82.7& \textbf{93.5} &87.8&82.6&83.1&82.9\\
        RoBERTa-base&76.9&75.4&76.2&86.1&89.3&87.7&88.9&87.0&87.9&84.0&83.9&83.9\\
        RoBERTa-large+SCL&78.4&73.1&75.7&86.9&87.2&87.0&\textbf{89.1}&89.1&89.1&\textbf{84.8}&83.1&83.9\\
        FT+POS & \textbf{78.9} &78.0&78.4&86.5&93.4&89.8&87.5&91.3&\textbf{89.4}&82.9&85.7&84.3\\
        Disc-ST &78.0& 82.4&80.1&89.9&90.8&90.4&79.6&93.5&86.0&83.6& 86.7 &85.2\\
        \hline
        FF2 - w/o TNP &76.3 & 81.9 & 79.0 & 89.3 &90.8 &90.0 &79.6 & 93.4 & 85.9 & 82.4 & 86.5 & 84.4 \\
        FF2 - w/o Interaction & 78.2 & 83.6 & 80.8 & 89.7 & 89.5 & 89.6 & 78.1 & 93.5 & 85.1 & 83.5 & 86.7 & 85.0 \\
        FF2   & 78.8 & \textbf{82.7} & \textbf{80.7}&89.3&90.7&90.0&79.6& \textbf{93.5} & 86.0 & 83.8 & \textbf{86.8} & \textbf{85.3} \\
        \bottomrule
  \end{tabular}
  }
\end{table*}

\subsection{FF2}
As mentioned, we design a two-stream architecture. One stream (ITE) is based on the pre-trained language model Electra \footnote{\url{https://huggingface.co/google/electra-large-discriminator}}, and another tiny non-pretrained module (TNP) is randomly initialized. The TNP is is significantly smaller than ITE. Here, we replace self-attention with Interaction to encourage message sharing.

During the training, we first obtain two feature vectors which are $C_i$ and $C_m$ via ITE and TNP modules. Second, $C_i$ and $C_m$ are concatenated as the input $H \in \mathbb{R}^{n \times (d_i+d_m)}$ of Fusion Layer (FL). FL is just one layer transformer, and the vector $H$ is fed into the FL to obtain the final output sequence $\hat{Y}$. Thus, we use cross-entropy loss $\ell$ as follows:
\begin{equation}
    \ell = - \sum_{i=1}^{K} p_i\log p_i
\end{equation}
Here, K is the total number of categories, $p_i$ is the predicted probability of label $i$.

\begin{table*}[!ht]
  \caption{Results in terms of P(\%), R(\%), F1(\%) on the English IWSLT2011 ASR transcription test set.}
  \label{tab:2}
  \centering
  \setlength{\tabcolsep}{2.5mm}{
    \begin{tabular}{c|r|r|r|r|r|r|r|r|r|r|r|r}
       \toprule
        \multirow{2}*{\textbf{Model}} & \multicolumn{3}{c|}{\textbf{Comma}} &\multicolumn{3}{c|}{\textbf{Period}} &\multicolumn{3}{c|}{\textbf{Question}} &\multicolumn{3}{c}{\textbf{Overall}} \\
        \cline{2-13} & P & R& F1 & P & R& F1 & P & R& F1 & P & R& F1\\
        \hline
        Self-attention-word-speech&64.0&59.6&61.7&75.5&75.5&75.6&\textbf{72.6}&65.9&\textbf{69.1}&70.7&67.1& - \\[0pt]
        BERT-base+Adversarial&\textbf{70.7}&68.1&\textbf{69.4}&77.3&77.5&77.5&68.4&66.0&67.2&\textbf{72.2}&70.5& - \\[0pt]
        BERT-base+FocalLoss&59.0&\textbf{76.6}&66.7&78.7&79.9&79.3&60.5&71.5&65.6&66.1&76.0&70.7 \\[0pt]
        RoBERTa-large+augmentation&64.1&68.8&66.3&81.0 &83.7&82.3&55.3&74.3&63.4&72.0&76.2&\textbf{74.0}\\[0pt]
        FT+POS&56.6&71.6&63.2&79.0&\textbf{87.0}&82.8&60.5&74.3&66.7&66.9&79.3&72.6\\[0pt]
        FF2 & 54.8&75.8&63.6&\textbf{82.5}&84.3&\textbf{83.4}&51.8&\textbf{82.9}&63.7 & 66.3& \textbf{80.1}&72.6\\[0pt]
        \bottomrule
          \end{tabular}
  }
\end{table*}

\section{Experiments}
In this section, we compare the performance of FF2 with the state-of-the-art approaches on the popular dataset in the fair comparison setting and further ablate some design choices in FF2 to understand their contributions.
\subsection{Dataset}
The English IWSLT2011\cite{IWSLT2012} is a popular benchmark dataset for punctuation restoration. It contains 2.1M words for training set, 296K words for validation set, and 12626 words for the manual transcription test set. It also contains 12822 words for the actual ASR transcription test set, in which the words are predicted by ASR systems, so it comprises some grammatical errors or wrong words. There are four types of punctuation marks (none, comma, period, and question mark), the distribution of categories in the training dataset is as follows: 85.7\% without punctuation, 7.53\% with a comma, 6.3\% with the period, and 0.47\% with question marks. In the following, we will employ precision (P), recall (R), and F1-score (F1) to evaluate FF2 and other approaches.

\subsection{Baselines}
\begin{sloppypar}
    We compare FF2 with the top-performing approaches. One part is RNN-based and employs a deep recurrent neural network, such as BLSTM-CRF, Teacher-Ensemble, and DRNN-LWMA-pre. The other line is transformer-based, including Self-attention-word-speech, CT-Transformer, and SAPR. SAPR uses a transformer encoder-decoder architecture and views punctuation restoration as a translation task. BERT-base+Adversarial, BERT-large+Transfer, BERT-base+FocalLoss (it utilizes the focal loss not cross-entropy loss to advance the results). RoBERTa-large+augmentation and RoBERTa-large+SCL. RoBERTa-base predicts the tags by multiple context windows, funnel-transformer-xlarge+POS+Fusion+SBS abbreviated as FT+POS incorporates an external POS tagger and fuses its predicted labels into the existing language model to address the problem. ELECTRA-large+Disc-ST named Disc-ST, which introduces a discriminative self-training approach with weighted loss and utilizes the extra dataset to achieve the SOTA performance.\cite{Yi2017DistillingKF,8682418,8682260,DBLP:journals/corr/abs-2003-01309,8545470,Adversarial,9023200,Yi2020,alam-etal-2020-punctuation,courtland-etal-2020-efficient,Shi2021IncorporatingEP,DBLP:journals/corr/abs-2107-09099, Discriminative}.
\end{sloppypar}

\subsection{Experiment Setup}
For the two-stream architecture, the Interaction Transformer Encoder is based on the public pre-trained language model (Electra), while tiny non-pretrained (TNP is 6 layers transformer blocks) and Fusion Layer are randomly initialized. Adam optimizer is set to default parameters. The learning rate of Electra and TNP are both 5e-6. Turning to the Fusion Layer, we use 8 attention heads, and the hidden size is 3072. Specifically, the maximum sequence length is 256, the batch size is 8, and the dropout rate in our experiments is set to 0.2. We also introduce R-Drop \cite{DBLP:journals/corr/abs-2106-14448} to act as a regularizer. Moreover, we adopt early stopping with patience of 8 epochs to avoid overfitting. We train and evaluate all the models on the 32GB Tesla V100.

\subsection{Overall Results}
Table \ref{tab:1} presents the results of FF2 compared to top performers. Apparently, the transformer-based approaches are far superior to those RNN-based. Our method accomplishes the best results in the Comma (recall, f1-score), the Question (recall), and the Overall (recall, f1-score), separately. Compared to Disc-ST and FT+POS, FF2 obtains a higher recall and f1-score,  and exceeds FT+POS by 1 point in the overall f1-score. Thus, the results validate that our proposed feature fusion two-stream framework is effective.

Some methods advance the performance by incorporating some extra information. For instance, Self-attention-word-speech utilizes both lexical and prosody features. FT+POS employs part-of-speech (POS for short) tools to add additional POS tags as extra knowledge. RoBERTa-large+augmentation construct the data by insertion, substitution, and deletion. Not like them, FF2 leverages the training dataset itself and focuses on the design of architecture. Without extra data, FF2 improves 1.0\%, 2.4\%, and 12.4\% compared to FT+POS, RoBERTa-large+augmentation, and Self-attention-word-speech, respectively.

In comparison with the current SOTA model Disc-ST, Disc-ST advances the performance by utilizing a self-training strategy and extending the training dataset (2M) with a large amount of unlabeled data (30M). Thus, their dataset is nearly 16 times larger than ours. As our model parameters are comparable to Disc-ST's, the results of FF2 (85.3) compared to Disc-ST (85.2) indicate that our design is robust.


For the results of the ASR transcription test set. Without focal loss or data augmentation, we can find that our approach outperforms the bert-based models by a large margin. Besides, FF2 obtains four best results in total, 82.5\% in precision and 83.4 \% in f1 of the period, 82.9\% in recall of question, and 80.1\% in  precision of the overall, respectively. Compared to RoBERTa-large+augmentation, FF2 has a slightly lower precision and a competitive F1-score but leads to 3.9 points improvement in recall. One possible reason is that data augmentation for RoBERTa-large+augmentation can benefit the robustness.



We conduct the ablation study and the results are shown in table \ref{tab:1}. First, the metric has a large drop (-0.9\%) while eliminating the TNP module, which verifies that TNP can capture the feature at hand and benefit ITE module. Second, The metric would decrease 0.3\% when removing Interaction, which indicates the modification of self-attention is valid.


\section{Conclusions}
In this paper, we propose a novel feature fusion framework (FF2) based on a two-type architecture with different perspectives. One stream utilizes a pre-trained language model to capture the semantic feature of the sequence, and another tiny module is randomly initialized to capture the feature information on the current dataset. We also replace the computation of self-attention with Interaction to encourage message sharing among heads. The experiments on the benchmark IWSLT demonstrate that our method is more effective than previous works. Since we do not employ external knowledge or data augmentation, future work can leverage extra information to boost the inference performance further.

\bibliographystyle{IEEEbib}
\bibliography{strings,refs}

\end{document}